\documentclass[times,twocolumn,final]{elsarticle}

\usepackage{framed,multirow}

\usepackage{amsmath}

\DeclareMathOperator*{\argmin}{arg\,min}

\usepackage{amssymb}
\usepackage{latexsym}

\usepackage{url}
\usepackage{xcolor}
\definecolor{newcolor}{rgb}{.8,.349,.1}

\usepackage{graphicx,subfigure}

\usepackage{hyperref}

\usepackage[switch,pagewise]{lineno} 

\begin{document}


\begin{frontmatter}

\title{Metal Artifact Reduction in Cone-Beam X-Ray CT via Ray Profile Correction}%

\author{Sungsoo Ha}
\author{Klaus Mueller*}

\address {Computer Science Department, Stony Brook University, Stony Brook, NY 11794-2424
\vskip 0.2 cm
*Corresponding author: mueller@cs.stonybrook.edu}



\begin{abstract}
In computed tomography (CT), metal implants increase the inconsistencies between
the measured data and the linear attenuation assumption made by analytic CT reconstruction
algorithms. The inconsistencies give rise to dark and bright bands and streaks in the reconstructed image, collectively called metal artifacts. These artifacts make it difficult for radiologists to render
correct diagnostic decisions. We describe a data-driven metal artifact reduction (MAR) algorithm
for image-guided spine surgery that applies to scenarios in which a prior CT scan of the patient is
available.
We tested the proposed method with two clinical datasets that were both obtained
during spine surgery. Using the proposed method, we were not only able to remove the dark
and bright streaks caused by the implanted screws but we also recovered the anatomical structures
hidden by these artifacts. 
This results in an improved
capability of surgeons to confirm the correctness of the implanted pedicle screw placements. 
\end{abstract}

\begin{keyword}
Computed tomography reconstruction \sep low dose CT \sep cone-beam CT \sep metal artifacts.
\end{keyword}

\end{frontmatter}


\section{Introduction}
\label{sec:intro}
X-ray computed tomography (CT) is a leading cross-sectional imaging technique lauded for its high image resolution and rapid speed of acquisition. CT reconstruction algorithms based on the Radon transform reconstruct the internal structures of human bodies by utilizing the X-ray photon interactions with matter which are following the Beer-Lambert law \cite{avinash1988principles}. However, this procedure takes the assumption that the X-ray source is monochromatic. When it comes to (more realistic) polychromatic X-ray sources, low-energy photons are attenuated more easily than high-energy ones and, as a result, the X-ray beam is \textit{hardened} as it passes through the material, shifting the energy spectrum toward higher energies. This becomes a particular problem when CT scanning is conducted with patients bearing metallic implants, which have dramatically increased attenuation properties for lower energies. The selective photon absorption not only increases the amount of dose absorbed by the object, it also amplifies the X-ray beam's hardness. Moreover, the implanted metals can severely change photon direction (and energy). A failure to consider all of these kinds of non-linear behaviors in the X-ray photon interactions (also include Poisson noise, photon starvation, motion, partial volume effect and etc) results in various artifacts, for example, dark streaks along the lines of greatest attenuation \cite{brooks1976beam,hsieh2009computed}. The high pass filter used in Filtered-back projection (FBP) \cite{feldkamp1984practical} then further exaggerates the differences between adjacent detector elements where one element has received a hardened beam and the other has not. This unintended contrast produces bright streaks in other directions. As a consequence, due to these adverse mechanisms, metal artifacts obscure information about anatomical structures, making it difficult for radiologists to correctly interpret the affected CT images.

There have been extensive efforts in developing metal artifact reduction (MAR) algorithms to compensate the approximation errors caused by implanted metals or high density objects. These efforts can be largely divided into two types of approaches: iterative reconstruction and sinogram correction. The iterative reconstruction methods adapt existing X-ray CT systems by incorporating one or more of the following types of a-priori knowledge: (1) low-level information of the images to be reconstructed \cite{xue2009metal,duan2008metal,zhang2011new,mehranian20133d}, (2) the X-ray spectrum of the source \cite{de2001iterative}, (3) the attenuation functions of the base materials \cite{elbakri2003segmentation,srivastava2005simplified,abella2009new}, and (4) the composition of the metal components \cite{stayman2012model}. More recently proposed iterative algorithms attempt to reduce beam hardening effects without the need of any prior knowledge by decomposing the image to be reconstructed into low and high density components \cite{kyriakou2010empirical,jin2015model}. 

On the other hand, the sinogram correction methods aim to directly correct the \textit{metal shadow} regions in the projection data in which the corresponding rays have interacted with metal objects. One early approach replaces the corrupted data with their neighbors using linear \cite{kalender1987reduction} or high-order interpolation schemes \cite{bazalova2007correction,abdoli2010reduction,zhao2000x}. However, the interpolation-based MARs often suffer from loss of detail around the metal objects, and they also have high propensity to introduce new streak artifacts \cite{muller2009spurious}. To address the lack of structural information, Prell et al. \cite{prell2009novel} and Meyer et al. \cite{meyer2010normalized} attempted to build prior CT images by roughly segmenting the uncorrected or pre-corrected CT image into soft-tissue, air, and bone equivalent materials. This has been a promising idea and further efforts have emerged that  seek to produce a better prior image with the help  of advanced computer vision techniques \cite{li2014prior,karimi2012segmentation}. Recently, for example, Zhang et al. \cite{zhang2017convolutional} utilized a convolutional neural network to generate a more sophisticated prior and used it to correct the sinogram contaminated by metal artifacts.      

In this paper we present a new MAR method that also uses the general approach of correcting a contaminated sinogram by substituting corrupted data with cleaner data available in prior images. The synthesis process we propose is not unlike the one often used in image-guided surgery (IGS) \cite{grimson1999image}. These methods perform a real-time correlation of the operative field with a preoperative imaging dataset to show the precise location of a selected surgical instrument in the surrounding anatomic structures. To realize this, before the surgery, the patient undergoes a series of CT scans that reveal the soft tissues and bony structures. In our scenario, these preoperative CT images serve as prior images to help remove the metal artifacts that appear in intra- or post-surgery CT scans due to the implanted metal objects. Since such prior images have been acquired from the same patient, they will likely contain very similar internal structures, especially around metal implants. Furthermore, as these  regions are often at least partially surrounded by bone it is unlikely that they are markedly deformed during the surgery. Thus, to find surrogate values to replace unreliable data in the sinogram, we first search ray paths in the prior images that have very similar density profiles along the ray passing through the metal objects. Then, the best matched prior ray profiles are used to correct the ray paths profiles that are corrupted by metal artifacts. Finally, the unreliable data are replaced with the re-projections of corrected ray profiles.  We explored this general idea in \cite{ha2016metal} but this preliminary work was limited to 2D fan-beam CT geometry. In this paper, we generalize our method to 3D cone-beam CT geometry and also present a significantly refined and mature framework. 

In the following, Section \ref{sec:mar_2} describes the proposed method and its technical details. Then, in Section \ref{sec:mar_3},  we show metal artifact reduction results. Section \ref{sec:mar_5} concludes the paper with a discussion on future research directions for the proposed method.

\section{Methods} 
\label{sec:mar_2}
In the following we use spine surgery as an example where a prior patient scan is available and an immediate post-surgery scan is required to confirm the correctness of the placements of the implanted metals screws.

\begin{figure*}[!bt]
  \centering
  \includegraphics[width=\textwidth]{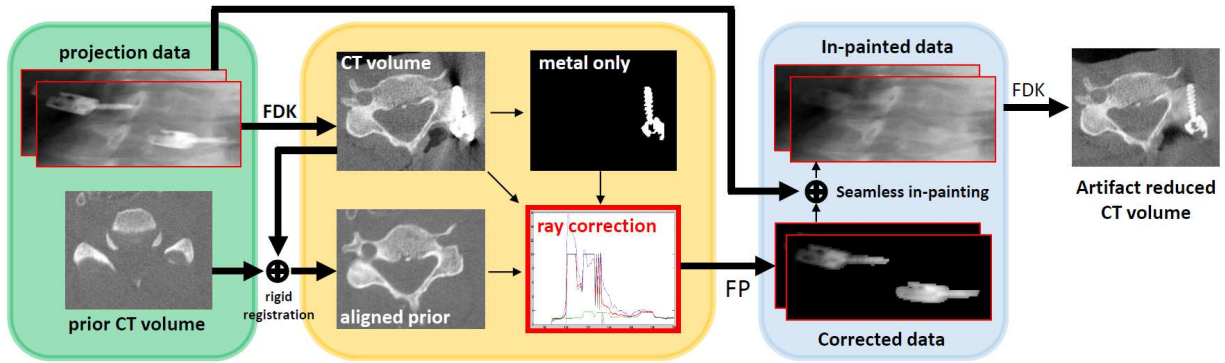}
  \caption{Overview of the proposed MAR method}
  \label{fig:mar_1}     
\end{figure*}

Our method requires two CT data sets (or one that has a sufficient portion free of metal). One data set is an artifact-free prior CT scan taken before  the (spine) surgery. The other is obtained during the surgery, containing metal artifacts due to the implanted pedicle screws. Although the two CT scans are taken from the same patient, they might be obtained in different conditions (e.g. patient’s pose, X-ray dosage amount, field-of-view and etc.). Therefore, it is necessary to register one volume to the other before applying the proposed ray profile correction method. This registration step is described in Section \ref{sec:mar_21}. Ray profile correction is also required to know which parts of a ray path belong to metal objects, and which ones do not. For this, we extract the implanted pedicle screws from the uncorrected CT volume (the volume suffering  from metal artifacts). This metal localization step is described in Section \ref{sec:mar_22}. In the following we shall call a set of sample points along a ray the \textit{ray profile}. A line integral is then computed as the weighted sum of all sample points of a ray profile. Also, we will define the regions traversed by rays passing through metal objects as \textit{metal shadow}. The projection values under the metal shadow are unreliable because of beam-hardening, photon starvation, and so on, and they will result in metal artifacts. Our goal is to compute surrogate values in the metal shadow regions by correcting the corresponding ray profiles using the aligned prior CT volume and geometric information of the implanted metal (here the pedicle screws). This new correction scheme is explained in Section \ref{sec:mar_23}. Finally, the corrected metal shadow is combined smoothly with the original CT projection data as described in Section \ref{sec:mar_24}. The overall process is illustrated in Figure \ref{fig:mar_1}. 


\subsection{Rigid volume registration}
\label{sec:mar_21}
For the ray profile correction, we need to find matched prior profiles from the prior CT volume generated in the pre-operative CT scan. This prior CT volume is usually significantly misaligned with the CT volume obtained during or after the surgery. The patient may be in a different pose or the CT scan may cover a different range of the spine region (or have a different field-of-view). All that these two volumes might have in common is the surgical region itself. One naive approach for finding matched ray profiles would be to exhaustively search the prior CT volume with the ray profiles extracted from the uncorrected CT volume. This approach would be computationally very demanding as the searching space is almost infinite. 

\begin{figure}[!bt]
  \centering
  \includegraphics[width=0.5\textwidth]{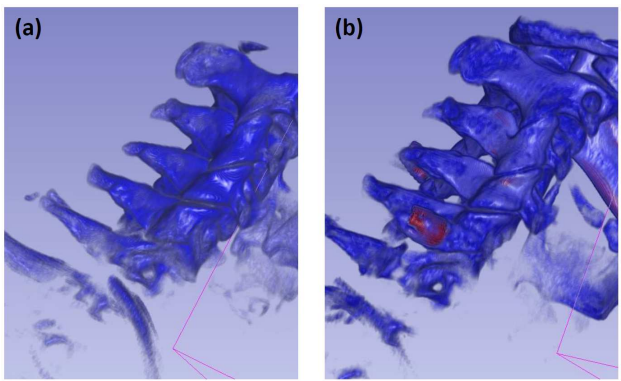}
  \caption{Similarity in bone structures between CT scans (a) before and (b) after a spine surgery. The visualizations are manually adjusted and colored to show the best view (Blue for bone and red for implanted pedicle screws and jaw that have higher density than bone).}
  \label{fig:mar_2}     
\end{figure}

\begin{figure}[!bt]
  \centering
  \includegraphics[width=0.5\textwidth]{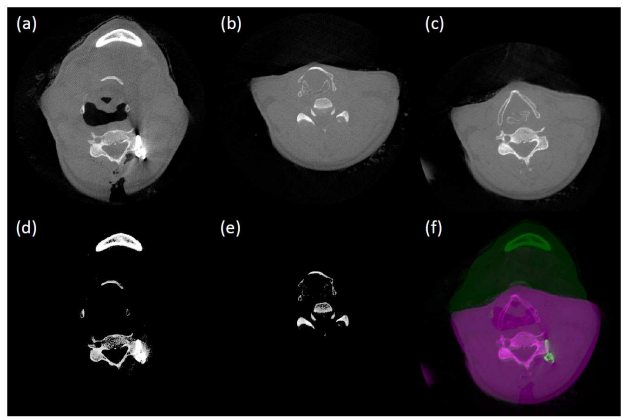}
  \caption{Rigid CT volume registration. (a) uncorrected CT volume, (b) prior CT volume, (c) registered prior CT volume to (a), (d) extracted bone structures of (a), (e) extracted bone structures of (b), (f) pseudo-colored overlap view of (a) and (c). Note that the presented CT slices, (a) and (b), initially show a large discrepancy although they have been sampled from the same z-index before applying the registration.}
  \label{fig:mar_4}     
\end{figure}

Instead, we align the two volumes and then find the set of matched prior profiles. The challenge in aligning two CT volumes taken at different times is that there can be large discrepancies in the soft internal structures (e.g. tissues). To resolve this problem, we first extract bone structures which are quite robust to deformation and also less contaminated by metal artifacts, in contrast to soft structures (see Figure \ref{fig:mar_2} for a visualization).  For the bone structure extraction, assuming there are only low and high density materials, we employ the balanced histogram thresholding (BHT) method \cite{dosbi}. Figure \ref{fig:mar_4} shows two CT volumes obtained before and after surgery along with bone structures extracted using the BHT method. The prior volume is then rigidly registered to the uncorrected one by solving the following minimization problem:
\begin{equation} \label{eq:mar_3}
\hat{\vec{\theta}} = \argmin_{ \vec{\theta} } \sum_{i=1}^{N} p_{i} \cdot |f_{i}^{unc} - T_{ \vec{\theta} }(f^{pri})_{i}| .
\end{equation}
Here, $f$ is the bone-only CT volume and its super-script $unc$ and $prior$ indicate the uncorrected and the prior CT volumes, respectively, which consist of $N$ voxels in total. The $T_{\vec{\theta}}(\cdot)$ is a rigid volume transformation operator with parameter vector, $\theta$, which includes $3$ translations and $3$ rotations in a 3D Cartesian coordinate system. Using Eq.(\ref{eq:mar_3}) we find the optimal parameter vector, $\hat{\vec{\theta}}$, which has a minimum weighted sum of the voxel-wise absolute difference between the two volumes. The weight term, $p_{i}$, is there to further penalize the data mismatch term if two voxels came from different anatomical structures (materials). It is formalized as follows: 
\begin{equation} \label{eq:mar_4}
p_{i} = \begin{cases} 
1 & \text{if } f_{i}^{unc} \text{ and } T_{\vec{\theta}}(f^{pri})_{i} \in \text{ same material} \\
c & \text{otherwise} \\
\end{cases}
\end{equation}
Here, we give more penalty, $c (\geq 1)$ if the two voxels are not in same material (i.e. two voxels are from bone structures or not). This minimizes the contribution of mis-categorized voxels in the uncorrected volume, such as voxels in the bright band or in implanted metals that are regarded as bone after applying the BHT segmentation.

We use a GPU-accelerated Hybrid-PSO (particle swarm optimization) algorithm to solve the minimization problem in Eq.(\ref{eq:mar_3}) which avoids a  convergence to a local minima \cite{kennedy2011particle,wachowiak2004approach}. More specifically, in each generation, we randomly choose half of the particles and randomly adjust either a translation or rotation parameter with uniform probability. In every third generation, we pick half of the worst particles. The first half of these are replaced with completely new random values. Among the remainder, three-fifth of the particles are randomized as we do in each generation and the crossover is applied to the others. These types of randomization strategies have proven effective in finding the global solution in different optimization tasks \cite{wachowiak2004approach,sharp2015accurate}. Figure \ref{fig:mar_4} shows an example of the rigid registration result.

\subsection{Localization of implanted metal objects}
\label{sec:mar_22}
\begin{figure}[!bt]
  \centering
  \includegraphics[width=0.48\textwidth]{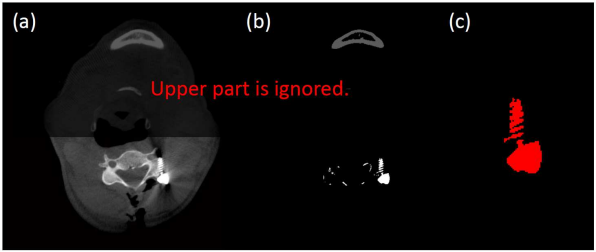}
  \caption{Metal localization. (a) uncorrected CT volume, (b) extracted high density materials using the BHT, and (c) final results with DBSCAN}
  \label{fig:mar_5}     
\end{figure}

Before we can correct the ray profiles (see Section \ref{sec:mar_23}), we need to know whether a sample point in a given ray profile originates from metal (or not). For this purpose, we segment the screws from the uncorrected CT volume as follow. In the first step, we use the balanced histogram thresholding (BHT) method \cite{dosbi} to get a coarse segmentation. After that we apply the DBSCAN (Density-based spatial clustering of applications with noise) algorithm \cite{ester1996density} to the segmented structures. DBSCAN  is a popular clustering algorithm which classifies points that are not well connected to a cluster as outliers.  We found that DBSCAN did very well to remove any remaining noise and obtain an accurate segmentation of the metal objects (in our case, the screws). The clean segmentation also allows us to to precisely determine  how many screws were implanted. Optionally, we might  also include prior geometric knowledge to accelerate the process and to further improve the clustering accuracy. Figure \ref{fig:mar_5} shows a screw extracted from an uncorrected CT volume. Figure \ref{fig:mar_5}c is what we will refer to as the \textit{metal-only CT volume}. 

\subsection{Ray profile correction}
\label{sec:mar_23}

To compute the profile's surrogate values (the values subject to replacement),  we use the observation that metal artifacts usually appear around implanted metals and that the degree of corruption tends to decrease with distance from the metal region. Using this observation, the noisy ray profiles are corrected by dividing it into two regions, metal and non-metal regions.  A metal region is the part of a profile that traverses a metal-only CT volume. For these regions, since we usually know the material of the implanted metals and their linear attenuation coefficients, the surrogate values are replaced with the linearly interpolated values of the two profiles extracted from the prior and the metal-only CT volume. We use linear interpolation to take into account the partial volume effect around the metal boundaries and so avoid any binarization artifacts. For the non-metal regions, the surrogate values are computed by linear interpolating between the noisy and prior profiles extracted from the uncorrected and the prior CT volume, respectively. The interpolation weight is given by the distance from the nearby metal boundaries along the ray path. It takes into account that when a point in a ray profile is close to a metal it is more likely deteriorated by metal artifacts and thus we put more emphasis on prior information. Vice versa, when a profile point is sufficiently far away from metal we can safely rely on its value in the uncorrected volume. As such, our method smoothly blends prior image information into the currently acquired imagery but only at locations where the current image information is likely unreliable due to metal artifacts. 

\begin{figure*}[!bt]
  \centering
  \includegraphics[width=\textwidth]{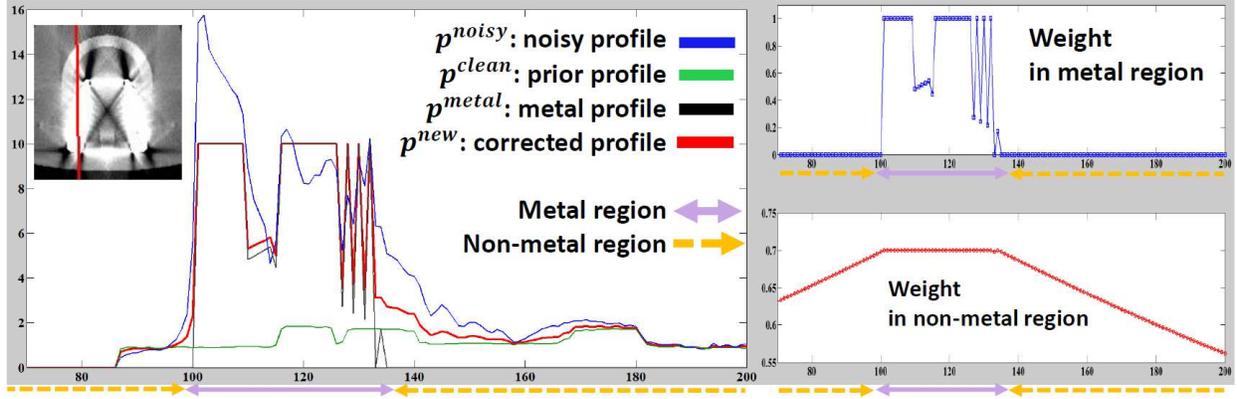}
  \caption{Ray profile correction. [left] ray profiles extracted from noisy, prior, metal-only and corrected CT volumes and interpolation weights in [right-top] metal region and [right-bottom] non-metal region. In these figures the x-axis denotes the sampled locations of the ray profiles, while the y-axis shows the intensity (or weight) values at these positions.}
  \label{fig:mar_6}     
\end{figure*}

Our ray profile correction scheme is described in Figure \ref{fig:mar_6} and can be formulated as follows:
\begin{equation} \label{eq:mar_5}
p_{i}^{new} = \begin{cases} 
lerp(\rho,p_{i}^{clean},p_{i}^{metal}/\rho) & \text{ if } i \in \text{ metal region } \\
lerp(p_{i}^{clean},p_{i}^{noisy},c \cdot 
\exp(-p_{i}^{dt}/h)) & \text{ if } i \in  \text{ non-metal region } \\
\end{cases}.
\end{equation}
where $lerp(\alpha,\beta,\omega)$ is the linear interpolation operator such that $\omega \cdot \alpha + (1-\omega) \cdot \beta$. In this equation,  $p_{i}$ represents the sampled value of a ray profile at position $i$ while its superscripts $new$, $metal$, $noisy$ and $clean$, indicate the corrected profile and the profiles extracted from the metal-only, uncorrected, and aligned prior volumes, respectively. The superscript, $dt$, denotes the distance transform of $p^{metal}$ and henceforth, $p_{i}^{dt}$  is the distance from the position $i$ to the closest metal boundary along the ray path \cite{maurer2003linear}. The value $\rho$ is the linear attenuation coefficient of the implanted metals while $h$ is a scalar that controls the smoothness of the weight factor and $c$ is a constant value representing how similar the aligned prior volume is to the uncorrected one. We experimentally determined it as $0.7$ in this work. 

\subsection{Seamless in-painting}
\label{sec:mar_24}
\begin{figure*}[!bt]
  \centering
  \includegraphics[width=0.6\textwidth]{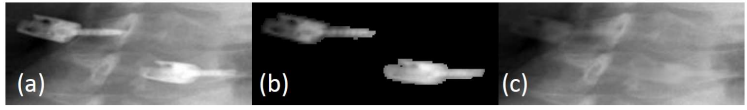}
  \caption{An example of seamless in-painting. (a) original, (b) corrected and (c) in-painted projection data}
  \label{fig:mar_7}     
\end{figure*}

Recall that sample profiles are only computed for rays that traverse the segmented metal in the metal-only CT volume. We integrate these rays and store them in a corrected sinogram.  Figure \ref{fig:mar_7}a shows a portion of an uncorrected sinogram while Figure \ref{fig:mar_7}b shows the same region with the corrected profiles only.  The final task is  to replace the  metal shadow regions of the uncorrected sinogram with these corrected regions. However, a direct replacement of the data can lead to undesired discontinuities around the boundary of the metal shadow, resulting in the generation of new artifacts \cite{muller2009spurious}. Therefore, it is important to seamlessly combine the new data with the existing ones at the boundary while internally keeping the relative contrast and the details of the data. To achieve such a seamless in-painting, we solve the following minimization problem \cite{perez2003poisson}:
\begin{equation}
\hat{P}^{new} = \min_{P^{new}} \sum_{i \in R} \left( \sum_{j \in N_{i} \cap R } (\nabla P_{j}^{new} - \nabla P_{j}^{corr} )^{2} + \sum_{j \in N_{i} \cap \sim R}  (P_{j}^{new} - P_{j}^{orig})^{2} \right)   ,\label{eq:mar_6} 
\end{equation}
where $P$ is projection data and its super-scripts, $new$, $corr$ and $orig$, represent the in-painted, corrected, and original projection data, respectively. $R$ denotes the metal shadow region and $N_{i}$  is 8-connected neighborhood of a pixel, $i$.  In this equation, the first term aims to preserve the gradients of the original (uncorrected) projection data in the metal shadow regions while lowering the intensities to non-metal values. Preserving the original gradients ensures that the detail and contrast in the projection data is maintained for the subsequent reconstruction.  The second term in the equation affects the metal shadow region's boundary only and ensures a smooth transition to the outside regions. Figure \ref{fig:mar_7}c shows a result of this seamless in-painting process using the same region than in panel a and b.

\section{Results and Discussions}
\label{sec:mar_3}
To test the proposed MAR framework, we used clinical CT projection data obtained during an image-guided surgery procedure on a  cervical spine region using a Medtronic O-Arm surgical imaging CT scanner. The scanner has a source to axis distance of $647.7$ $mm$  and a source to detector distance of $1147.7$ $mm$.  It is equipped with a  flat X-ray detector with $1024 \times 384$ bins and an active area of  $393.432 \times 290.224$ $mm^{2}$. During the scan, 360 projections  were collected uniformly distributed over 360$^{\circ}$. The 3D reconstruction used the filtered back-projection algorithm \cite{avinash1988principles} and produced a  $512 \times 512 \times 192$ volume with a voxel size of  $0.415 \times 0.415 \times 0.83$ $mm^{3}$.

\begin{figure*}[!bt]
  \centering
  \includegraphics[width=0.8\textwidth]{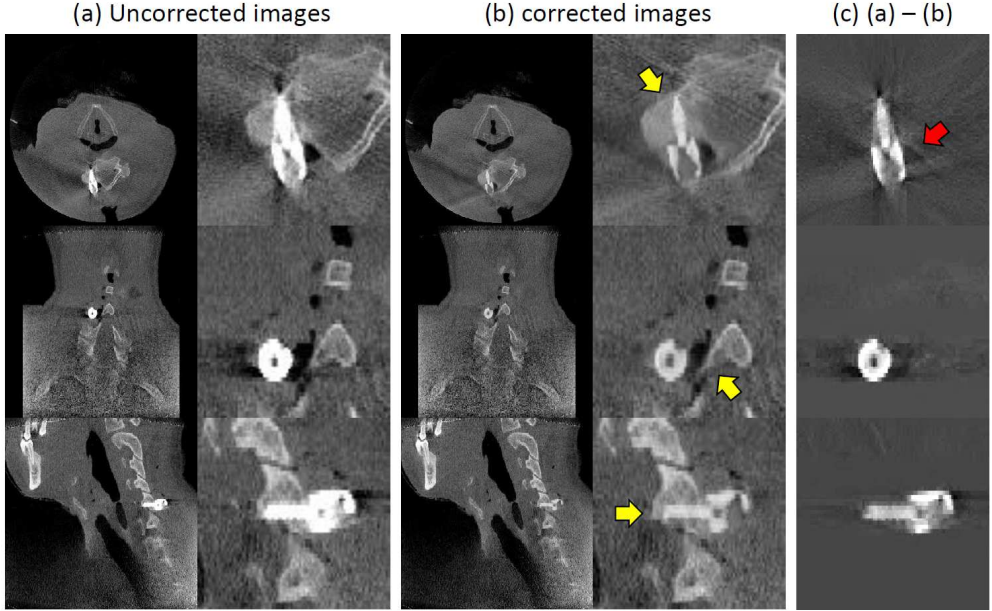}
  \caption{Case study I. From top to bottom, transverse, coronal and sigittal views.}
  \label{fig:case_b1}     
\end{figure*}

\begin{figure*}[h]
  \centering
  \includegraphics[width=0.8\textwidth]{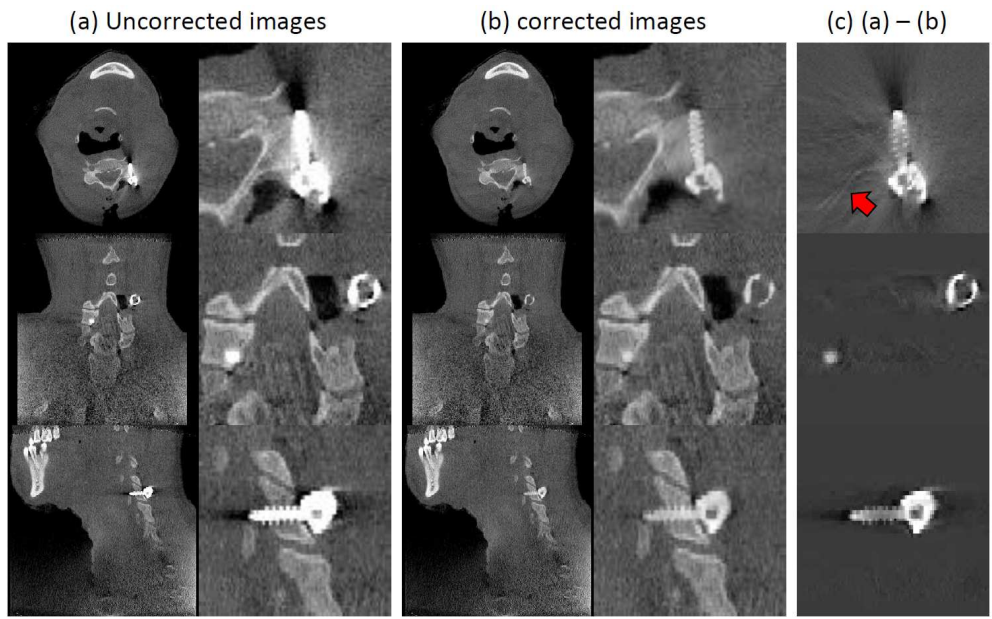}
  \caption{Case study II. From top to bottom, transverse, coronal and sagittal views.}
  \label{fig:case_b2}     
\end{figure*}

Figures \ref{fig:case_b1} and \ref{fig:case_b2} show some results we obtained using our metal artifact reduction algorithm. The spine has  two pedicle screws implanted. Figure \ref{fig:case_b1} shows one of them in transverse, sagittal and coronal views. Note that the sagittal and coronal views are horizontal and vertical cut slices passing through the screws, respectively. Figure \ref{fig:case_b2} shows the other implanted screws in the same manner. Overall, the proposed method effectively removes the metal artifacts (dark/bright bands and streaks) and reveals clear outlines of the implanted pedicle screws which are suitable for evaluating their placements after the surgery.  For example, in Figure \ref{fig:case_b1}, top  and bottom row, the yellow arrow indicates a pedicle screw where only the corrected image (column b) can reveal that is has been correctly inserted into the bone without extending into the tissue. Likewise, the yellow arrow in the middle row in Figure \ref{fig:case_b1}) shows a volume feature that was  previously hidden by the beam hardening artifacts (column a) but is now readily visible. 

One side effect of the proposed method is the tendency of blurring the anatomical structures near metal objects. The difference images (column c) between the uncorrected and corrected images show (1) the removed artifacts, (2) a bright version of the metal pieces (as mentioned, our method lowers their projection values in the sinogram), and (3) some incorrectly removed  details. The latter causes the blurring effects (annotated by the red arrow in Figure \ref{fig:case_b1} and Figure \ref{fig:case_b2}). We think this is primarily because of the distance-based artifact region prediction model in Eq.(\ref{eq:mar_5}) where the model estimates the artifact regions based on the distance ($p_{i}^{dt}$) from a point to the nearest metal boundary along a ray path regardless of whether the point is corrupted by metal artifacts or not. One way to mitigate the blurring effect could be to introduce an additional stage at the end of our MAR framework that would exploit the information hidden in the low- and high-pass filtered sinograms \cite{jeong2009metal} or images \cite{meyer2012frequency} to control the correction process. 

\section{Conclusion}
\label{sec:mar_5}
We have presented a new method for metal artifact reduction (MAR). It assumes that a prior  CT scan taken before implanting the metal objects into the patient is available. Using this prior scan and a segmentation or a model of the metal implant, we utilize a novel ray profile correction scheme which computes an accurate estimate of the rays traversing  the projection regions affected by the metal artifacts. Our experiments with clinical CT data indicate that the proposed method can clearly reveal the placements of implanted pedicle screws that were ambiguous before due  the significant metal artifacts.

Future work will extend this work along three directions.  Firstly, we will further investigate the behavior of the phenomenon of beam hardening for better metal artifact region prediction. Secondly, using a CAD model available for commercial pedicle screws, we believe a better localization of the implanted metal can be accomplished, leading to a better ray profile correction. Lastly, we also believe that utilizing existing algorithms that can reveal features  hidden by metal artifacts in conjunction with a MAR method will be able to help resolve the occasional blurring effects around metals. Finally, we also  plan to conduct  more clinical case studies which will contain a wide selection of different implanted metals and different amounts of metal artifacts.


%
%

%

\section*{Acknowledgement}
This research was partially supported by NSF grant IIS 1527200 and the Ministry of Science, ICT and Future Planning, Korea, under the “IT Consilience Creative Program (ITCCP)” supervised by NIPA. We also thank Medtronic, Inc. for the data and support. 

\section*{References}
\bibliographystyle{cag-num-names}
\bibliography{myReferences}

\begin{thebibliography}{36}%
\makeatletter
\providecommand \@ifxundefined [1]{%
 \@ifx{#1\undefined}
}%
\providecommand \@ifnum [1]{%
 \ifnum #1\expandafter \@firstoftwo
 \else \expandafter \@secondoftwo
 \fi
}%
\providecommand \@ifx [1]{%
 \ifx #1\expandafter \@firstoftwo
 \else \expandafter \@secondoftwo
 \fi
}%
\providecommand \natexlab [1]{#1}%
\providecommand \enquote  [1]{``#1''}%
\providecommand \bibnamefont  [1]{#1}%
\providecommand \bibfnamefont [1]{#1}%
\providecommand \citenamefont [1]{#1}%
\providecommand \href@noop [0]{\@secondoftwo}%
\providecommand \href [0]{\begingroup \@sanitize@url \@href}%
\providecommand \@href[1]{\@@startlink{#1}\@@href}%
\providecommand \@@href[1]{\endgroup#1\@@endlink}%
\providecommand \@sanitize@url [0]{\catcode `\\12\catcode `\$12\catcode
  `\&12\catcode `\#12\catcode `\^12\catcode `\_12\catcode `\%12\relax}%
\providecommand \@@startlink[1]{}%
\providecommand \@@endlink[0]{}%
\providecommand \url  [0]{\begingroup\@sanitize@url \@url }%
\providecommand \@url [1]{\endgroup\@href {#1}{\urlprefix }}%
\providecommand \urlprefix  [0]{URL }%
\providecommand \Eprint [0]{\href }%
\providecommand \doibase [0]{http://dx.doi.org/}%
\providecommand \selectlanguage [0]{\@gobble}%
\providecommand \bibinfo  [0]{\@secondoftwo}%
\providecommand \bibfield  [0]{\@secondoftwo}%
\providecommand \translation [1]{[#1]}%
\providecommand \BibitemOpen [0]{}%
\providecommand \bibitemStop [0]{}%
\providecommand \bibitemNoStop [0]{.\EOS\space}%
\providecommand \EOS [0]{\spacefactor3000\relax}%
\providecommand \BibitemShut  [1]{\csname bibitem#1\endcsname}%
\let\auto@bib@innerbib\@empty
\bibitem [{\citenamefont {Kak}\ and\ \citenamefont
  {Slaney}(1988)}]{avinash1988principles}%
  \BibitemOpen
  \bibfield  {author} {\bibinfo {author} {\bibfnamefont {Avinash~C..}\
  \bibnamefont {Kak}}\ and\ \bibinfo {author} {\bibfnamefont {Malcolm}\
  \bibnamefont {Slaney}},\ }\href@noop {} {\emph {\bibinfo {title} {Principles
  of computerized tomographic imaging}}}\ (\bibinfo  {publisher} {IEEE press},\
  \bibinfo {year} {1988})\BibitemShut {NoStop}%
\bibitem [{\citenamefont {Brooks}\ and\ \citenamefont
  {Di~Chiro}(1976)}]{brooks1976beam}%
  \BibitemOpen
  \bibfield  {author} {\bibinfo {author} {\bibfnamefont {Rodney~A}\
  \bibnamefont {Brooks}}\ and\ \bibinfo {author} {\bibfnamefont {Giovanni}\
  \bibnamefont {Di~Chiro}},\ }\bibfield  {title} {\enquote {\bibinfo {title}
  {Beam hardening in x-ray reconstructive tomography},}\ }\href@noop {}
  {\bibfield  {journal} {\bibinfo  {journal} {Physics in medicine and biology}\
  }\textbf {\bibinfo {volume} {21}},\ \bibinfo {pages} {390} (\bibinfo {year}
  {1976})}\BibitemShut {NoStop}%
\bibitem [{\citenamefont {Hsieh}(2009)}]{hsieh2009computed}%
  \BibitemOpen
  \bibfield  {author} {\bibinfo {author} {\bibfnamefont {Jiang}\ \bibnamefont
  {Hsieh}},\ }\bibfield  {title} {\enquote {\bibinfo {title} {Computed
  tomography: principles, design, artifacts, and recent advances},}\ \
  }(\bibinfo {organization} {SPIE Bellingham, WA},\ \bibinfo {year}
  {2009})\BibitemShut {NoStop}%
\bibitem [{\citenamefont {Feldkamp}\ \emph {et~al.}(1984)\citenamefont
  {Feldkamp}, \citenamefont {Davis},\ and\ \citenamefont
  {Kress}}]{feldkamp1984practical}%
  \BibitemOpen
  \bibfield  {author} {\bibinfo {author} {\bibfnamefont {LA}~\bibnamefont
  {Feldkamp}}, \bibinfo {author} {\bibfnamefont {LC}~\bibnamefont {Davis}}, \
  and\ \bibinfo {author} {\bibfnamefont {JW}~\bibnamefont {Kress}},\ }\bibfield
   {title} {\enquote {\bibinfo {title} {Practical cone-beam algorithm},}\
  }\href@noop {} {\bibfield  {journal} {\bibinfo  {journal} {JOSA A}\ }\textbf
  {\bibinfo {volume} {1}},\ \bibinfo {pages} {612--619} (\bibinfo {year}
  {1984})}\BibitemShut {NoStop}%
\bibitem [{\citenamefont {Xue}\ \emph {et~al.}(2009)\citenamefont {Xue},
  \citenamefont {Zhang}, \citenamefont {Xiao}, \citenamefont {Chen},\ and\
  \citenamefont {Xing}}]{xue2009metal}%
  \BibitemOpen
  \bibfield  {author} {\bibinfo {author} {\bibfnamefont {Hui}\ \bibnamefont
  {Xue}}, \bibinfo {author} {\bibfnamefont {Li}~\bibnamefont {Zhang}}, \bibinfo
  {author} {\bibfnamefont {Yongshun}\ \bibnamefont {Xiao}}, \bibinfo {author}
  {\bibfnamefont {Zhiqiang}\ \bibnamefont {Chen}}, \ and\ \bibinfo {author}
  {\bibfnamefont {Yuxiang}\ \bibnamefont {Xing}},\ }\bibfield  {title}
  {\enquote {\bibinfo {title} {Metal artifact reduction in dual energy ct by
  sinogram segmentation based on active contour model and tv inpainting},}\
  }in\ \href@noop {} {\emph {\bibinfo {booktitle} {2009 IEEE Nuclear Science
  Symposium Conference Record (NSS/MIC)}}}\ (\bibinfo {organization} {IEEE},\
  \bibinfo {year} {2009})\ pp.\ \bibinfo {pages} {904--908}\BibitemShut
  {NoStop}%
\bibitem [{\citenamefont {Duan}\ \emph {et~al.}(2008)\citenamefont {Duan},
  \citenamefont {Zhang}, \citenamefont {Xiao}, \citenamefont {Cheng},
  \citenamefont {Chen},\ and\ \citenamefont {Xing}}]{duan2008metal}%
  \BibitemOpen
  \bibfield  {author} {\bibinfo {author} {\bibfnamefont {Xinhui}\ \bibnamefont
  {Duan}}, \bibinfo {author} {\bibfnamefont {Li}~\bibnamefont {Zhang}},
  \bibinfo {author} {\bibfnamefont {Yongshun}\ \bibnamefont {Xiao}}, \bibinfo
  {author} {\bibfnamefont {Jianping}\ \bibnamefont {Cheng}}, \bibinfo {author}
  {\bibfnamefont {Zhiqiang}\ \bibnamefont {Chen}}, \ and\ \bibinfo {author}
  {\bibfnamefont {Yuxiang}\ \bibnamefont {Xing}},\ }\bibfield  {title}
  {\enquote {\bibinfo {title} {Metal artifact reduction in ct images by
  sinogram tv inpainting},}\ }in\ \href@noop {} {\emph {\bibinfo {booktitle}
  {2008 IEEE Nuclear Science Symposium Conference Record}}}\ (\bibinfo
  {organization} {IEEE},\ \bibinfo {year} {2008})\ pp.\ \bibinfo {pages}
  {4175--4177}\BibitemShut {NoStop}%
\bibitem [{\citenamefont {Zhang}\ \emph {et~al.}(2011)\citenamefont {Zhang},
  \citenamefont {Pu}, \citenamefont {Hu}, \citenamefont {Liu},\ and\
  \citenamefont {Zhou}}]{zhang2011new}%
  \BibitemOpen
  \bibfield  {author} {\bibinfo {author} {\bibfnamefont {Yi}~\bibnamefont
  {Zhang}}, \bibinfo {author} {\bibfnamefont {Yi-Fei}\ \bibnamefont {Pu}},
  \bibinfo {author} {\bibfnamefont {Jin-Rong}\ \bibnamefont {Hu}}, \bibinfo
  {author} {\bibfnamefont {Yan}\ \bibnamefont {Liu}}, \ and\ \bibinfo {author}
  {\bibfnamefont {Ji-Liu}\ \bibnamefont {Zhou}},\ }\bibfield  {title} {\enquote
  {\bibinfo {title} {A new ct metal artifacts reduction algorithm based on
  fractional-order sinogram inpainting},}\ }\href@noop {} {\bibfield  {journal}
  {\bibinfo  {journal} {Journal of X-ray science and technology}\ }\textbf
  {\bibinfo {volume} {19}},\ \bibinfo {pages} {373--384} (\bibinfo {year}
  {2011})}\BibitemShut {NoStop}%
\bibitem [{\citenamefont {Mehranian}\ \emph {et~al.}(2013)\citenamefont
  {Mehranian}, \citenamefont {Ay}, \citenamefont {Rahmim},\ and\ \citenamefont
  {Zaidi}}]{mehranian20133d}%
  \BibitemOpen
  \bibfield  {author} {\bibinfo {author} {\bibfnamefont {Abolfazl}\
  \bibnamefont {Mehranian}}, \bibinfo {author} {\bibfnamefont {Mohammad~Reza}\
  \bibnamefont {Ay}}, \bibinfo {author} {\bibfnamefont {Arman}\ \bibnamefont
  {Rahmim}}, \ and\ \bibinfo {author} {\bibfnamefont {Habib}\ \bibnamefont
  {Zaidi}},\ }\bibfield  {title} {\enquote {\bibinfo {title} {3d prior image
  constrained projection completion for x-ray ct metal artifact reduction},}\
  }\href@noop {} {\bibfield  {journal} {\bibinfo  {journal} {IEEE Transactions
  on Nuclear Science}\ }\textbf {\bibinfo {volume} {60}},\ \bibinfo {pages}
  {3318--3332} (\bibinfo {year} {2013})}\BibitemShut {NoStop}%
\bibitem [{\citenamefont {De~Man}\ \emph {et~al.}(2001)\citenamefont {De~Man},
  \citenamefont {Nuyts}, \citenamefont {Dupont}, \citenamefont {Marchal},\ and\
  \citenamefont {Suetens}}]{de2001iterative}%
  \BibitemOpen
  \bibfield  {author} {\bibinfo {author} {\bibfnamefont {Bruno}\ \bibnamefont
  {De~Man}}, \bibinfo {author} {\bibfnamefont {Johan}\ \bibnamefont {Nuyts}},
  \bibinfo {author} {\bibfnamefont {Patrick}\ \bibnamefont {Dupont}}, \bibinfo
  {author} {\bibfnamefont {Guy}\ \bibnamefont {Marchal}}, \ and\ \bibinfo
  {author} {\bibfnamefont {Paul}\ \bibnamefont {Suetens}},\ }\bibfield  {title}
  {\enquote {\bibinfo {title} {An iterative maximum-likelihood polychromatic
  algorithm for ct},}\ }\href@noop {} {\bibfield  {journal} {\bibinfo
  {journal} {IEEE transactions on medical imaging}\ }\textbf {\bibinfo {volume}
  {20}},\ \bibinfo {pages} {999--1008} (\bibinfo {year} {2001})}\BibitemShut
  {NoStop}%
\bibitem [{\citenamefont {Elbakri}\ and\ \citenamefont
  {Fessler}(2003)}]{elbakri2003segmentation}%
  \BibitemOpen
  \bibfield  {author} {\bibinfo {author} {\bibfnamefont {Idris~A}\ \bibnamefont
  {Elbakri}}\ and\ \bibinfo {author} {\bibfnamefont {Jeffrey~A}\ \bibnamefont
  {Fessler}},\ }\bibfield  {title} {\enquote {\bibinfo {title}
  {Segmentation-free statistical image reconstruction for polyenergetic x-ray
  computed tomography with experimental validation},}\ }\href@noop {}
  {\bibfield  {journal} {\bibinfo  {journal} {Physics in medicine and biology}\
  }\textbf {\bibinfo {volume} {48}},\ \bibinfo {pages} {2453} (\bibinfo {year}
  {2003})}\BibitemShut {NoStop}%
\bibitem [{\citenamefont {Srivastava}\ and\ \citenamefont
  {Fessler}(2005)}]{srivastava2005simplified}%
  \BibitemOpen
  \bibfield  {author} {\bibinfo {author} {\bibfnamefont {Somesh}\ \bibnamefont
  {Srivastava}}\ and\ \bibinfo {author} {\bibfnamefont {Jeffrey~A}\
  \bibnamefont {Fessler}},\ }\bibfield  {title} {\enquote {\bibinfo {title}
  {Simplified statistical image reconstruction algorithm for polyenergetic
  x-ray ct},}\ }in\ \href@noop {} {\emph {\bibinfo {booktitle} {IEEE Nuclear
  Science Symposium Conference Record}}},\ Vol.~\bibinfo {volume} {3}\
  (\bibinfo {organization} {Citeseer},\ \bibinfo {year} {2005})\ pp.\ \bibinfo
  {pages} {1551--1555}\BibitemShut {NoStop}%
\bibitem [{\citenamefont {Abella}\ and\ \citenamefont
  {Fessler}(2009)}]{abella2009new}%
  \BibitemOpen
  \bibfield  {author} {\bibinfo {author} {\bibfnamefont {M{\'o}nica}\
  \bibnamefont {Abella}}\ and\ \bibinfo {author} {\bibfnamefont {Jeffrey~A}\
  \bibnamefont {Fessler}},\ }\bibfield  {title} {\enquote {\bibinfo {title} {A
  new statistical image reconstruction algorithm for polyenergetic x-ray ct},}\
  }in\ \href@noop {} {\emph {\bibinfo {booktitle} {2009 IEEE International
  Symposium on Biomedical Imaging: From Nano to Macro}}}\ (\bibinfo
  {organization} {IEEE},\ \bibinfo {year} {2009})\ pp.\ \bibinfo {pages}
  {165--168}\BibitemShut {NoStop}%
\bibitem [{\citenamefont {Stayman}\ \emph {et~al.}(2012)\citenamefont
  {Stayman}, \citenamefont {Otake}, \citenamefont {Prince}, \citenamefont
  {Khanna},\ and\ \citenamefont {Siewerdsen}}]{stayman2012model}%
  \BibitemOpen
  \bibfield  {author} {\bibinfo {author} {\bibfnamefont {J~Webster}\
  \bibnamefont {Stayman}}, \bibinfo {author} {\bibfnamefont {Yoshito}\
  \bibnamefont {Otake}}, \bibinfo {author} {\bibfnamefont {Jerry~L}\
  \bibnamefont {Prince}}, \bibinfo {author} {\bibfnamefont {A~Jay}\
  \bibnamefont {Khanna}}, \ and\ \bibinfo {author} {\bibfnamefont {Jeffrey~H}\
  \bibnamefont {Siewerdsen}},\ }\bibfield  {title} {\enquote {\bibinfo {title}
  {Model-based tomographic reconstruction of objects containing known
  components},}\ }\href@noop {} {\bibfield  {journal} {\bibinfo  {journal}
  {IEEE transactions on medical imaging}\ }\textbf {\bibinfo {volume} {31}},\
  \bibinfo {pages} {1837--1848} (\bibinfo {year} {2012})}\BibitemShut {NoStop}%
\bibitem [{\citenamefont {Kyriakou}\ \emph {et~al.}(2010)\citenamefont
  {Kyriakou}, \citenamefont {Meyer}, \citenamefont {Prell},\ and\ \citenamefont
  {Kachelrie{\ss}}}]{kyriakou2010empirical}%
  \BibitemOpen
  \bibfield  {author} {\bibinfo {author} {\bibfnamefont {Yiannis}\ \bibnamefont
  {Kyriakou}}, \bibinfo {author} {\bibfnamefont {Esther}\ \bibnamefont
  {Meyer}}, \bibinfo {author} {\bibfnamefont {Daniel}\ \bibnamefont {Prell}}, \
  and\ \bibinfo {author} {\bibfnamefont {Marc}\ \bibnamefont
  {Kachelrie{\ss}}},\ }\bibfield  {title} {\enquote {\bibinfo {title}
  {Empirical beam hardening correction (ebhc) for ct},}\ }\href@noop {}
  {\bibfield  {journal} {\bibinfo  {journal} {Medical physics}\ }\textbf
  {\bibinfo {volume} {37}},\ \bibinfo {pages} {5179--5187} (\bibinfo {year}
  {2010})}\BibitemShut {NoStop}%
\bibitem [{\citenamefont {Jin}\ \emph {et~al.}(2015)\citenamefont {Jin},
  \citenamefont {Bouman},\ and\ \citenamefont {Sauer}}]{jin2015model}%
  \BibitemOpen
  \bibfield  {author} {\bibinfo {author} {\bibfnamefont {Pengchong}\
  \bibnamefont {Jin}}, \bibinfo {author} {\bibfnamefont {Charles~A}\
  \bibnamefont {Bouman}}, \ and\ \bibinfo {author} {\bibfnamefont {Ken~D}\
  \bibnamefont {Sauer}},\ }\bibfield  {title} {\enquote {\bibinfo {title} {A
  model-based image reconstruction algorithm with simultaneous beam hardening
  correction for x-ray ct},}\ }\href@noop {} {\bibfield  {journal} {\bibinfo
  {journal} {IEEE Transactions on Computational Imaging}\ }\textbf {\bibinfo
  {volume} {1}},\ \bibinfo {pages} {200--216} (\bibinfo {year}
  {2015})}\BibitemShut {NoStop}%
\bibitem [{\citenamefont {Kalender}\ \emph {et~al.}(1987)\citenamefont
  {Kalender}, \citenamefont {Hebel},\ and\ \citenamefont
  {Ebersberger}}]{kalender1987reduction}%
  \BibitemOpen
  \bibfield  {author} {\bibinfo {author} {\bibfnamefont {Willi~A}\ \bibnamefont
  {Kalender}}, \bibinfo {author} {\bibfnamefont {Robert}\ \bibnamefont
  {Hebel}}, \ and\ \bibinfo {author} {\bibfnamefont {Johannes}\ \bibnamefont
  {Ebersberger}},\ }\bibfield  {title} {\enquote {\bibinfo {title} {Reduction
  of ct artifacts caused by metallic implants.}}\ }\href@noop {} {\bibfield
  {journal} {\bibinfo  {journal} {Radiology}\ }\textbf {\bibinfo {volume}
  {164}},\ \bibinfo {pages} {576--577} (\bibinfo {year} {1987})}\BibitemShut
  {NoStop}%
\bibitem [{\citenamefont {Bazalova}\ \emph {et~al.}(2007)\citenamefont
  {Bazalova}, \citenamefont {Beaulieu}, \citenamefont {Palefsky},\ and\
  \citenamefont {Verhaegen}}]{bazalova2007correction}%
  \BibitemOpen
  \bibfield  {author} {\bibinfo {author} {\bibfnamefont {Magdalena}\
  \bibnamefont {Bazalova}}, \bibinfo {author} {\bibfnamefont {Luc}\
  \bibnamefont {Beaulieu}}, \bibinfo {author} {\bibfnamefont {Steven}\
  \bibnamefont {Palefsky}}, \ and\ \bibinfo {author} {\bibfnamefont {Frank}\
  \bibnamefont {Verhaegen}},\ }\bibfield  {title} {\enquote {\bibinfo {title}
  {Correction of ct artifacts and its influence on monte carlo dose
  calculations},}\ }\href@noop {} {\bibfield  {journal} {\bibinfo  {journal}
  {Medical physics}\ }\textbf {\bibinfo {volume} {34}},\ \bibinfo {pages}
  {2119--2132} (\bibinfo {year} {2007})}\BibitemShut {NoStop}%
\bibitem [{\citenamefont {Abdoli}\ \emph {et~al.}(2010)\citenamefont {Abdoli},
  \citenamefont {Ay}, \citenamefont {Ahmadian}, \citenamefont {Dierckx},\ and\
  \citenamefont {Zaidi}}]{abdoli2010reduction}%
  \BibitemOpen
  \bibfield  {author} {\bibinfo {author} {\bibfnamefont {Mehrsima}\
  \bibnamefont {Abdoli}}, \bibinfo {author} {\bibfnamefont {Mohammad~Reza}\
  \bibnamefont {Ay}}, \bibinfo {author} {\bibfnamefont {Alireza}\ \bibnamefont
  {Ahmadian}}, \bibinfo {author} {\bibfnamefont {Rudi~AJO}\ \bibnamefont
  {Dierckx}}, \ and\ \bibinfo {author} {\bibfnamefont {Habib}\ \bibnamefont
  {Zaidi}},\ }\bibfield  {title} {\enquote {\bibinfo {title} {Reduction of
  dental filling metallic artifacts in ct-based attenuation correction of pet
  data using weighted virtual sinograms optimized by a genetic algorithm},}\
  }\href@noop {} {\bibfield  {journal} {\bibinfo  {journal} {Medical physics}\
  }\textbf {\bibinfo {volume} {37}},\ \bibinfo {pages} {6166--6177} (\bibinfo
  {year} {2010})}\BibitemShut {NoStop}%
\bibitem [{\citenamefont {Zhao}\ \emph {et~al.}(2000)\citenamefont {Zhao},
  \citenamefont {Robeltson}, \citenamefont {Wang}, \citenamefont {Whiting},\
  and\ \citenamefont {Bae}}]{zhao2000x}%
  \BibitemOpen
  \bibfield  {author} {\bibinfo {author} {\bibfnamefont {Shiying}\ \bibnamefont
  {Zhao}}, \bibinfo {author} {\bibfnamefont {DD}~\bibnamefont {Robeltson}},
  \bibinfo {author} {\bibfnamefont {Ge}~\bibnamefont {Wang}}, \bibinfo {author}
  {\bibfnamefont {Bruce}\ \bibnamefont {Whiting}}, \ and\ \bibinfo {author}
  {\bibfnamefont {Kyongtae~T}\ \bibnamefont {Bae}},\ }\bibfield  {title}
  {\enquote {\bibinfo {title} {X-ray ct metal artifact reduction using
  wavelets: an application for imaging total hip prostheses},}\ }\href@noop {}
  {\bibfield  {journal} {\bibinfo  {journal} {IEEE transactions on medical
  imaging}\ }\textbf {\bibinfo {volume} {19}},\ \bibinfo {pages} {1238--1247}
  (\bibinfo {year} {2000})}\BibitemShut {NoStop}%
\bibitem [{\citenamefont {M{\"u}ller}\ and\ \citenamefont
  {Buzug}(2009)}]{muller2009spurious}%
  \BibitemOpen
  \bibfield  {author} {\bibinfo {author} {\bibfnamefont {Jan}\ \bibnamefont
  {M{\"u}ller}}\ and\ \bibinfo {author} {\bibfnamefont {Thorsten~M}\
  \bibnamefont {Buzug}},\ }\bibfield  {title} {\enquote {\bibinfo {title}
  {Spurious structures created by interpolation-based ct metal artifact
  reduction},}\ }in\ \href@noop {} {\emph {\bibinfo {booktitle} {SPIE Medical
  Imaging}}}\ (\bibinfo {organization} {International Society for Optics and
  Photonics},\ \bibinfo {year} {2009})\ pp.\ \bibinfo {pages}
  {72581Y--72581Y}\BibitemShut {NoStop}%
\bibitem [{\citenamefont {Prell}\ \emph {et~al.}(2009)\citenamefont {Prell},
  \citenamefont {Kyriakou}, \citenamefont {Beister},\ and\ \citenamefont
  {Kalender}}]{prell2009novel}%
  \BibitemOpen
  \bibfield  {author} {\bibinfo {author} {\bibfnamefont {Daniel}\ \bibnamefont
  {Prell}}, \bibinfo {author} {\bibfnamefont {Yiannis}\ \bibnamefont
  {Kyriakou}}, \bibinfo {author} {\bibfnamefont {Marcel}\ \bibnamefont
  {Beister}}, \ and\ \bibinfo {author} {\bibfnamefont {Willi~A}\ \bibnamefont
  {Kalender}},\ }\bibfield  {title} {\enquote {\bibinfo {title} {A novel
  forward projection-based metal artifact reduction method for flat-detector
  computed tomography},}\ }\href@noop {} {\bibfield  {journal} {\bibinfo
  {journal} {Physics in medicine and biology}\ }\textbf {\bibinfo {volume}
  {54}},\ \bibinfo {pages} {6575} (\bibinfo {year} {2009})}\BibitemShut
  {NoStop}%
\bibitem [{\citenamefont {Meyer}\ \emph {et~al.}(2010)\citenamefont {Meyer},
  \citenamefont {Raupach}, \citenamefont {Lell}, \citenamefont {Schmidt},\ and\
  \citenamefont {Kachelrie{\ss}}}]{meyer2010normalized}%
  \BibitemOpen
  \bibfield  {author} {\bibinfo {author} {\bibfnamefont {Esther}\ \bibnamefont
  {Meyer}}, \bibinfo {author} {\bibfnamefont {Rainer}\ \bibnamefont {Raupach}},
  \bibinfo {author} {\bibfnamefont {Michael}\ \bibnamefont {Lell}}, \bibinfo
  {author} {\bibfnamefont {Bernhard}\ \bibnamefont {Schmidt}}, \ and\ \bibinfo
  {author} {\bibfnamefont {Marc}\ \bibnamefont {Kachelrie{\ss}}},\ }\bibfield
  {title} {\enquote {\bibinfo {title} {Normalized metal artifact reduction
  (nmar) in computed tomography},}\ }\href@noop {} {\bibfield  {journal}
  {\bibinfo  {journal} {Medical physics}\ }\textbf {\bibinfo {volume} {37}},\
  \bibinfo {pages} {5482--5493} (\bibinfo {year} {2010})}\BibitemShut {NoStop}%
\bibitem [{\citenamefont {Li}\ \emph {et~al.}(2014)\citenamefont {Li},
  \citenamefont {Liu}, \citenamefont {Dong}, \citenamefont {Su},\ and\
  \citenamefont {Luo}}]{li2014prior}%
  \BibitemOpen
  \bibfield  {author} {\bibinfo {author} {\bibfnamefont {Ming}\ \bibnamefont
  {Li}}, \bibinfo {author} {\bibfnamefont {Zhaobang}\ \bibnamefont {Liu}},
  \bibinfo {author} {\bibfnamefont {Yuefang}\ \bibnamefont {Dong}}, \bibinfo
  {author} {\bibfnamefont {Kai}\ \bibnamefont {Su}}, \ and\ \bibinfo {author}
  {\bibfnamefont {Kangxin}\ \bibnamefont {Luo}},\ }\bibfield  {title} {\enquote
  {\bibinfo {title} {A prior-interpolation based metal artifact reduction
  algorithm in computed tomography},}\ }in\ \href@noop {} {\emph {\bibinfo
  {booktitle} {2014 7th International Conference on Biomedical Engineering and
  Informatics}}}\ (\bibinfo {organization} {IEEE},\ \bibinfo {year} {2014})\
  pp.\ \bibinfo {pages} {24--28}\BibitemShut {NoStop}%
\bibitem [{\citenamefont {Karimi}\ \emph {et~al.}(2012)\citenamefont {Karimi},
  \citenamefont {Cosman}, \citenamefont {Wald},\ and\ \citenamefont
  {Martz}}]{karimi2012segmentation}%
  \BibitemOpen
  \bibfield  {author} {\bibinfo {author} {\bibfnamefont {Seemeen}\ \bibnamefont
  {Karimi}}, \bibinfo {author} {\bibfnamefont {Pamela}\ \bibnamefont {Cosman}},
  \bibinfo {author} {\bibfnamefont {Christoph}\ \bibnamefont {Wald}}, \ and\
  \bibinfo {author} {\bibfnamefont {Harry}\ \bibnamefont {Martz}},\ }\bibfield
  {title} {\enquote {\bibinfo {title} {Segmentation of artifacts and anatomy in
  ct metal artifact reduction},}\ }\href@noop {} {\bibfield  {journal}
  {\bibinfo  {journal} {Medical physics}\ }\textbf {\bibinfo {volume} {39}},\
  \bibinfo {pages} {5857--5868} (\bibinfo {year} {2012})}\BibitemShut {NoStop}%
\bibitem [{\citenamefont {Zhang}\ and\ \citenamefont
  {Yu}(2017)}]{zhang2017convolutional}%
  \BibitemOpen
  \bibfield  {author} {\bibinfo {author} {\bibfnamefont {Yanbo}\ \bibnamefont
  {Zhang}}\ and\ \bibinfo {author} {\bibfnamefont {Hengyong}\ \bibnamefont
  {Yu}},\ }\bibfield  {title} {\enquote {\bibinfo {title} {Convolutional neural
  network based metal artifact reduction in x-ray computed tomography},}\
  }\href@noop {} {\bibfield  {journal} {\bibinfo  {journal} {arXiv preprint
  arXiv:1709.01581}\ } (\bibinfo {year} {2017})}\BibitemShut {NoStop}%
\bibitem [{\citenamefont {Grimson}\ \emph {et~al.}(1999)\citenamefont
  {Grimson}, \citenamefont {Kikinis}, \citenamefont {Jolesz},\ and\
  \citenamefont {Black}}]{grimson1999image}%
  \BibitemOpen
  \bibfield  {author} {\bibinfo {author} {\bibfnamefont {WEL}\ \bibnamefont
  {Grimson}}, \bibinfo {author} {\bibfnamefont {RJFA}\ \bibnamefont {Kikinis}},
  \bibinfo {author} {\bibfnamefont {Ferenc~A}\ \bibnamefont {Jolesz}}, \ and\
  \bibinfo {author} {\bibfnamefont {PM}~\bibnamefont {Black}},\ }\bibfield
  {title} {\enquote {\bibinfo {title} {Image-guided surgery},}\ }\href@noop {}
  {\bibfield  {journal} {\bibinfo  {journal} {Scientific American}\ }\textbf
  {\bibinfo {volume} {280}},\ \bibinfo {pages} {54--61} (\bibinfo {year}
  {1999})}\BibitemShut {NoStop}%
\bibitem [{\citenamefont {Ha}\ and\ \citenamefont
  {Mueller}(2016)}]{ha2016metal}%
  \BibitemOpen
  \bibfield  {author} {\bibinfo {author} {\bibfnamefont {Sungsoo}\ \bibnamefont
  {Ha}}\ and\ \bibinfo {author} {\bibfnamefont {Klaus}\ \bibnamefont
  {Mueller}},\ }\bibfield  {title} {\enquote {\bibinfo {title} {Metal artifact
  reduction in ct via ray profile correction},}\ }in\ \href@noop {} {\emph
  {\bibinfo {booktitle} {SPIE Medical Imaging}}}\ (\bibinfo {organization}
  {International Society for Optics and Photonics},\ \bibinfo {year} {2016})\
  pp.\ \bibinfo {pages} {978334--978334}\BibitemShut {NoStop}%
\bibitem [{\citenamefont {Anjos}\ and\ \citenamefont
  {Shahbazkia}(2008)}]{dosbi}%
  \BibitemOpen
  \bibfield  {author} {\bibinfo {author} {\bibfnamefont {Ant{\'o}nio}\
  \bibnamefont {Anjos}}\ and\ \bibinfo {author} {\bibfnamefont {Hamid~Reza}\
  \bibnamefont {Shahbazkia}},\ }\bibfield  {title} {\enquote {\bibinfo {title}
  {Bi-level image thresholding - a fast method},}\ }\href@noop {} {\bibfield
  {journal} {\bibinfo  {journal} {BIOSIGNALS}\ }\textbf {\bibinfo {volume}
  {2}},\ \bibinfo {pages} {70--76} (\bibinfo {year} {2008})}\BibitemShut
  {NoStop}%
\bibitem [{\citenamefont {Kennedy}(2011)}]{kennedy2011particle}%
  \BibitemOpen
  \bibfield  {author} {\bibinfo {author} {\bibfnamefont {James}\ \bibnamefont
  {Kennedy}},\ }\bibfield  {title} {\enquote {\bibinfo {title} {Particle swarm
  optimization},}\ }in\ \href@noop {} {\emph {\bibinfo {booktitle}
  {Encyclopedia of machine learning}}}\ (\bibinfo  {publisher} {Springer},\
  \bibinfo {year} {2011})\ pp.\ \bibinfo {pages} {760--766}\BibitemShut
  {NoStop}%
\bibitem [{\citenamefont {Wachowiak}\ \emph {et~al.}(2004)\citenamefont
  {Wachowiak}, \citenamefont {Smol{\'\i}kov{\'a}}, \citenamefont {Zheng},
  \citenamefont {Zurada},\ and\ \citenamefont
  {Elmaghraby}}]{wachowiak2004approach}%
  \BibitemOpen
  \bibfield  {author} {\bibinfo {author} {\bibfnamefont {Mark~P}\ \bibnamefont
  {Wachowiak}}, \bibinfo {author} {\bibfnamefont {Renata}\ \bibnamefont
  {Smol{\'\i}kov{\'a}}}, \bibinfo {author} {\bibfnamefont {Yufeng}\
  \bibnamefont {Zheng}}, \bibinfo {author} {\bibfnamefont {Jacek~M}\
  \bibnamefont {Zurada}}, \ and\ \bibinfo {author} {\bibfnamefont {Adel~Said}\
  \bibnamefont {Elmaghraby}},\ }\bibfield  {title} {\enquote {\bibinfo {title}
  {An approach to multimodal biomedical image registration utilizing particle
  swarm optimization},}\ }\href@noop {} {\bibfield  {journal} {\bibinfo
  {journal} {IEEE Transactions on evolutionary computation}\ }\textbf {\bibinfo
  {volume} {8}},\ \bibinfo {pages} {289--301} (\bibinfo {year}
  {2004})}\BibitemShut {NoStop}%
\bibitem [{\citenamefont {Sharp}\ \emph {et~al.}(2015)\citenamefont {Sharp},
  \citenamefont {Keskin}, \citenamefont {Robertson}, \citenamefont {Taylor},
  \citenamefont {Shotton}, \citenamefont {Kim}, \citenamefont {Rhemann},
  \citenamefont {Leichter}, \citenamefont {Vinnikov}, \citenamefont {Wei} \emph
  {et~al.}}]{sharp2015accurate}%
  \BibitemOpen
  \bibfield  {author} {\bibinfo {author} {\bibfnamefont {Toby}\ \bibnamefont
  {Sharp}}, \bibinfo {author} {\bibfnamefont {Cem}\ \bibnamefont {Keskin}},
  \bibinfo {author} {\bibfnamefont {Duncan}\ \bibnamefont {Robertson}},
  \bibinfo {author} {\bibfnamefont {Jonathan}\ \bibnamefont {Taylor}}, \bibinfo
  {author} {\bibfnamefont {Jamie}\ \bibnamefont {Shotton}}, \bibinfo {author}
  {\bibfnamefont {David}\ \bibnamefont {Kim}}, \bibinfo {author} {\bibfnamefont
  {Christoph}\ \bibnamefont {Rhemann}}, \bibinfo {author} {\bibfnamefont {Ido}\
  \bibnamefont {Leichter}}, \bibinfo {author} {\bibfnamefont {Alon}\
  \bibnamefont {Vinnikov}}, \bibinfo {author} {\bibfnamefont {Yichen}\
  \bibnamefont {Wei}},  \emph {et~al.},\ }\bibfield  {title} {\enquote
  {\bibinfo {title} {Accurate, robust, and flexible real-time hand tracking},}\
  }in\ \href@noop {} {\emph {\bibinfo {booktitle} {Proceedings of the 33rd
  Annual ACM Conference on Human Factors in Computing Systems}}}\ (\bibinfo
  {organization} {ACM},\ \bibinfo {year} {2015})\ pp.\ \bibinfo {pages}
  {3633--3642}\BibitemShut {NoStop}%
\bibitem [{\citenamefont {Ester}\ \emph {et~al.}(1996)\citenamefont {Ester},
  \citenamefont {Kriegel}, \citenamefont {Sander}, \citenamefont {Xu} \emph
  {et~al.}}]{ester1996density}%
  \BibitemOpen
  \bibfield  {author} {\bibinfo {author} {\bibfnamefont {Martin}\ \bibnamefont
  {Ester}}, \bibinfo {author} {\bibfnamefont {Hans-Peter}\ \bibnamefont
  {Kriegel}}, \bibinfo {author} {\bibfnamefont {J{\"o}rg}\ \bibnamefont
  {Sander}}, \bibinfo {author} {\bibfnamefont {Xiaowei}\ \bibnamefont {Xu}},
  \emph {et~al.},\ }\bibfield  {title} {\enquote {\bibinfo {title} {A
  density-based algorithm for discovering clusters in large spatial databases
  with noise.}}\ }in\ \href@noop {} {\emph {\bibinfo {booktitle} {Kdd}}},\
  Vol.~\bibinfo {volume} {96}\ (\bibinfo {year} {1996})\ pp.\ \bibinfo {pages}
  {226--231}\BibitemShut {NoStop}%
\bibitem [{\citenamefont {Maurer}\ \emph {et~al.}(2003)\citenamefont {Maurer},
  \citenamefont {Qi},\ and\ \citenamefont {Raghavan}}]{maurer2003linear}%
  \BibitemOpen
  \bibfield  {author} {\bibinfo {author} {\bibfnamefont {Calvin~R}\
  \bibnamefont {Maurer}}, \bibinfo {author} {\bibfnamefont {Rensheng}\
  \bibnamefont {Qi}}, \ and\ \bibinfo {author} {\bibfnamefont {Vijay}\
  \bibnamefont {Raghavan}},\ }\bibfield  {title} {\enquote {\bibinfo {title} {A
  linear time algorithm for computing exact euclidean distance transforms of
  binary images in arbitrary dimensions},}\ }\href@noop {} {\bibfield
  {journal} {\bibinfo  {journal} {IEEE Transactions on Pattern Analysis and
  Machine Intelligence}\ }\textbf {\bibinfo {volume} {25}},\ \bibinfo {pages}
  {265--270} (\bibinfo {year} {2003})}\BibitemShut {NoStop}%
\bibitem [{\citenamefont {P{\'e}rez}\ \emph {et~al.}(2003)\citenamefont
  {P{\'e}rez}, \citenamefont {Gangnet},\ and\ \citenamefont
  {Blake}}]{perez2003poisson}%
  \BibitemOpen
  \bibfield  {author} {\bibinfo {author} {\bibfnamefont {Patrick}\ \bibnamefont
  {P{\'e}rez}}, \bibinfo {author} {\bibfnamefont {Michel}\ \bibnamefont
  {Gangnet}}, \ and\ \bibinfo {author} {\bibfnamefont {Andrew}\ \bibnamefont
  {Blake}},\ }\bibfield  {title} {\enquote {\bibinfo {title} {Poisson image
  editing},}\ }in\ \href@noop {} {\emph {\bibinfo {booktitle} {ACM Transactions
  on graphics (TOG)}}},\ Vol.~\bibinfo {volume} {22}\ (\bibinfo {organization}
  {ACM},\ \bibinfo {year} {2003})\ pp.\ \bibinfo {pages} {313--318}\BibitemShut
  {NoStop}%
\bibitem [{\citenamefont {Jeong}\ and\ \citenamefont
  {Ra}(2009)}]{jeong2009metal}%
  \BibitemOpen
  \bibfield  {author} {\bibinfo {author} {\bibfnamefont {Kye~Young}\
  \bibnamefont {Jeong}}\ and\ \bibinfo {author} {\bibfnamefont {Jong~Beom}\
  \bibnamefont {Ra}},\ }\bibfield  {title} {\enquote {\bibinfo {title} {Metal
  artifact reduction based on sinogram correction in ct},}\ }in\ \href@noop {}
  {\emph {\bibinfo {booktitle} {2009 IEEE Nuclear Science Symposium Conference
  Record (NSS/MIC)}}}\ (\bibinfo {organization} {IEEE},\ \bibinfo {year}
  {2009})\ pp.\ \bibinfo {pages} {3480--3483}\BibitemShut {NoStop}%
\bibitem [{\citenamefont {Meyer}\ \emph {et~al.}(2012)\citenamefont {Meyer},
  \citenamefont {Raupach}, \citenamefont {Lell}, \citenamefont {Schmidt},\ and\
  \citenamefont {Kachelrie{\ss}}}]{meyer2012frequency}%
  \BibitemOpen
  \bibfield  {author} {\bibinfo {author} {\bibfnamefont {Esther}\ \bibnamefont
  {Meyer}}, \bibinfo {author} {\bibfnamefont {Rainer}\ \bibnamefont {Raupach}},
  \bibinfo {author} {\bibfnamefont {Michael}\ \bibnamefont {Lell}}, \bibinfo
  {author} {\bibfnamefont {Bernhard}\ \bibnamefont {Schmidt}}, \ and\ \bibinfo
  {author} {\bibfnamefont {Marc}\ \bibnamefont {Kachelrie{\ss}}},\ }\bibfield
  {title} {\enquote {\bibinfo {title} {Frequency split metal artifact reduction
  (fsmar) in computed tomography},}\ }\href@noop {} {\bibfield  {journal}
  {\bibinfo  {journal} {Medical physics}\ }\textbf {\bibinfo {volume} {39}},\
  \bibinfo {pages} {1904--1916} (\bibinfo {year} {2012})}\BibitemShut {NoStop}%
\end{thebibliography}

\end{document}